\documentclass[10pt, journal, final]{IEEEtran}

\ifCLASSINFOpdf
  \usepackage[pdftex]{graphicx}
\fi
\usepackage{amsmath}

\usepackage{amsfonts}
\usepackage{algorithm}
\usepackage{algpseudocode}
\usepackage{multirow}
\usepackage{hyperref}
\usepackage{arydshln}
\usepackage{xcolor}

\begin{document}

\title{Pruning with Compensation: Efficient Channel Pruning for Deep Convolutional Neural Networks}
\author{
Zhouyang Xie, Yan Fu, Shengzhao Tian, Junlin Zhou, Duanbing Chen
\thanks{Z. Xie, Y. Fu, J. Zhou and D. Chen(corresponding author, email: dbchen@uestc.edu.cn) are with Chengdu Union Big Data Technology Co. Ltd., Chengdu 610094, China.}
\thanks{S. Tian is with School of Computer Science and Engineering, University of Electronic Science and Technology of China, Chengdu 611731, China.}
\thanks{Y. Fu, J. Zhou and D. Chen are also with School of Computer Science and Engineering, University of Electronic Science and Technology of China.}
\thanks{Source code for this paper is available at \url{https://github.com/ZhouyangXie/compensation}}
}
\markboth{To be submitted to IEEE transactions on neural networks and learning systems}%
{Xie \MakeLowercase{\textit{et al.}}: Pruning with Compensation: Efficient Channel Pruning for Deep Convolutional Neural Networks}

\maketitle
\begin{abstract}

Channel pruning is a promising technique to compress the parameters of deep convolutional neural networks(DCNN) and to speed up the inference. This paper aims to address the long-standing inefficiency of channel pruning. Most channel pruning methods recover the prediction accuracy by re-training the pruned model from the remaining parameters or random initialization. This re-training process is heavily dependent on the sufficiency of computational resources, training data, and human interference(tuning the training strategy). In this paper, a highly efficient pruning method is proposed to significantly reduce the cost of pruning DCNN. The main contributions of our method include: 1) pruning compensation, a fast and data-efficient substitute of re-training to minimize the post-pruning reconstruction loss of features, 2) compensation-aware pruning(CaP), a novel pruning algorithm to remove redundant or less-weighted channels by minimizing the loss of information, and 3) binary structural search with step constraint to minimize human interference. On benchmarks including CIFAR-10/100 and ImageNet, our method shows competitive pruning performance among the state-of-the-art retraining-based pruning methods and, more importantly, reduces the processing time by $95\%$ and data usage by $90\%$.

\end{abstract}


\IEEEpeerreviewmaketitle

\section{Introduction}
\label{sec:introduction}

The success of deep convolutional neural networks(DCNN) in a wide range of visual tasks is achieved at the price of massive computational resource and high power consumption. The cost of DCNN inference is the major obstacle to deployment on edge devices and real-time services. Model compression for DCNN is the summary of techniques to reduce the computational workload of DCNN at various granularity. The compressed model is expected to have accelerated inference, reduced power consumption, and wider hardware applicability. Model compression techniques include the design of lightweight models(e.g. \cite{zhang2018shufflenet, forrest2016squzzenet, sandler2018mobilenetv2}), neural architecture search(e.g. \cite{zoph2016neural, baker2017accelerating, cai2018efficient, stanley2019designing}), knowledge distillation(e.g. \cite{hinton2015distilling,chen2017learning,park2019relational}), pruning(e.g. \cite{hassibi1993optimal, molchanov2016pruning,he2017channel}), and quantization(e.g. \cite{hubara2016binarized,jacob2018quantization,micikevicius2017mixed}) \textit{etc}. Therein, pruning is a technique to reduce the computational workload of DCNN through the removal of parameters(weights) in convolutional or fully-connected layers.  

\begin{figure}[!t]
    \centering
    \includegraphics[width=3.2in]{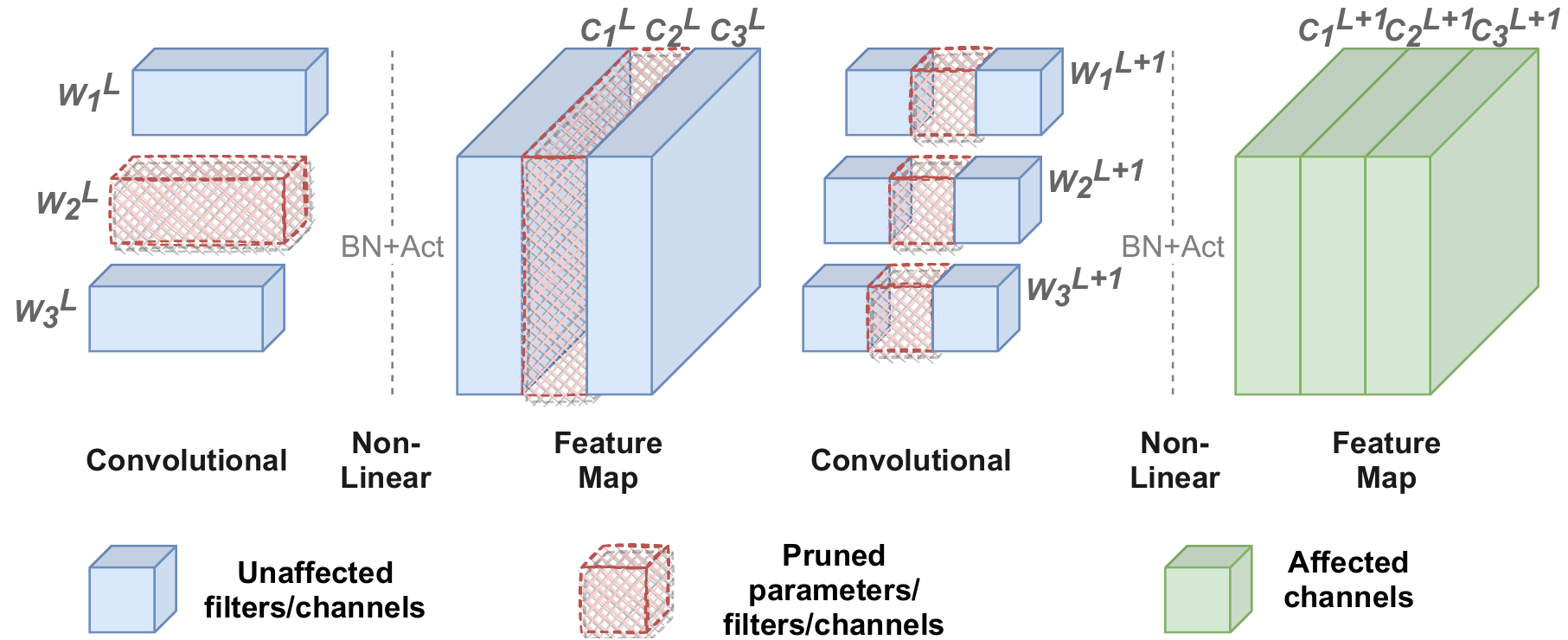}
    \caption{An illustration of channel pruning. In this example, the input channel $C_2^L$ of the $(L+1)$-th convolutional layer is pruned(red), so are the dependent weights in $W_2^{L+1}$ and filter $W_2^L$. Other weights are unaffected(blue). But the output of the $(L+1)$-th layer are all affected(green) because of the missing channel. The gray dashed line(denoted by \textit{BN+Act} in the figure) represents batch normalization and non-linear activation functions, also unaffected.}
    \label{fig:pruning_illustration}
\end{figure}

This paper focuses on structural pruning of DCNN, which removes kernels(also called filters) integrally. In contrast, un-structural pruning removes some of the parameters in a kernel and results in an incomplete kernel, which is incompatible with modern software/hardware implementation of convolution operation. Because structural pruning is performed at filter-level or channel-level(equivalent in practice), it's also called filter pruning or channel pruning. Figure \ref{fig:pruning_illustration} illustrates the removal of parameters when channel pruning takes effect.

A pruning method is designed under the trade-off between prediction accuracy and computational cost. Pruning will inevitably damage the original inference of the pre-trained model and often incur drop in prediction accuracy. Pruning methods are evaluated by their ability to reduce computational cost and preserve prediction accuracy simultaneously.

The development of channel/filter pruning methodology is centered on the strategy to preserve important channels/filters and remove unimportant channels/filters in a pre-trained model. It involves structural search, the search for a proper sparsity rate of each layer, and channel/filter selection, the method to make channel/filter removal and preservation decisions. Sparsity rate is the ratio of channels/filters to be removed in a layer. Higher sparsity will lead to less computational cost but more uncertainty to model inference. Because the structure of DCNN is determined by the sparsity rate of each layer, this step is also called structural search. Meanwhile, a ready pruning method takes accuracy recovery into account. Accuracy recovery is the process of regaining prediction accuracy of model after pruning. The most widely accepted approach is re-training the pruned model based on the remaining parameters and the data used for pre-training.

Despite channel pruning has shown promising compression performance on many DCNN architectures and visual tasks, the efficiency of channel pruning is a long-standing problem that hinders wide application, fast evaluation of method, and further exploration. More specifically, accuracy recovery by re-training the pruned model demands a long training process, considerable training data, and laborious human interference, while the computation cost of channel selection and structural search(without evaluation) are relatively negligible. We enumerate the cost of re-training as below:

\begin{itemize}
    \item Computational resource: re-training a model requires considerable GPU hours. For example, on ImageNet\cite{russakovsky2015imagenet}, a popular benchmark for visual recognition, re-training a pruned model usually demands more than 100 epochs and consumes nearly 100 GPU hours. On small benchmarks such as CIFAR\cite{krizhevsky2009cifar}, re-training by tens of epochs demands 5-6 GPU hours. Note that a pruning pipeline often requires many iterations to evaluate different structures, so the time consumption of re-training is multiplied.
    \item Data dependency: by default, the training data used in pre-training is also fully exploited by re-training. This dependency will raise concern when pruning models pre-trained on private or limitedly accessible data.
    \item Human interference: the strategy for re-training, including the setting of learning rate, regularization, training epoch, and optimization algorithm \textit{etc}, requires engineering experience as well as labour-costly human interference. How to design a proper re-training strategy for a pruned model is still an open question, as we review in section \ref{sec:related_works}.
\end{itemize}

This paper aims to improve the efficiency of channel pruning on DCNN by reducing above three cost items. 

We propose a novel and efficient accuracy recovery method, named as pruning compensation, to replace re-training. We are motivated by the observation that, if the pruning algorithm removes the redundant channels and keeps the diversity of the representation, the layer output can be recovered by fitting the missing signal with the remaining ones. This idea is feasible to convolution, because convolution is fundamentally a linear function and the compensation can be implemented as an addition to the linear transform. This implementation manner does not modify the structure of the pruned model and cause zero computational overhead at inference time.

Pruning compensation is expected to be more efficient than re-training. This one-step optimization of parameters is faster and more stable than gradient-based iterative optimization of re-training. Pruning compensation needs a reasonable amount of data to estimate the distribution of features(layer input) so that the remaining channels can fit the pruned channel. Experimental results show that the data dependency of pruning compensation is significant lower than that of re-training. What's more, pruning compensation needs no human interference.

Though pruning compensation as an accuracy recovery method is decoupled with pruning algorithm, namely any pruning algorithm is compatible with pruning compensation, we still find that existing pruning algorithms are not suitable for pruning compensation. Theoretically, the optimization objective of existing pruning algorithms is not aligned to the objective of pruning compensation, as formulated in subsection \ref{subsec:pruning_compensation}. We need to design a novel pruning algorithm that unifies the weight and the redundancy of channels. Thus, we propose compensation-aware pruning(CaP) in this paper to fully leverage the effectiveness of pruning compensation. It features the unification of channel weight and redundancy, and an efficient greedy strategy to solve the complex combinatorial problem of channel selection. Experiments in section \ref{sec:experiments} also exhibit the advantage of CaP over other existing pruning algorithms.

To completely automate the pruning process, we propose structural search with step constraint to find optimal sparsity settings. Thanks to the efficiency of pruning compensation, we can evaluate a structure faster than re-training and thus extend the search space. Therefore, we propose to search for an optimal sparsity rate layer-by-layer by binary search under a user-specified constraint. We add a step constraint to control the search process and prevent over-pruning early visited layers. In this pipeline, users only need to set a tolerance value to constrain the accuracy drop of the pruned model, and no other human interference is involved.

To summarize our contributions, we propose an efficient channel pruning pipeline in this paper. The efficiency is mainly improved by pruning compensation, a fast, data-efficient and interference-free accuracy recovery method that replaces the inefficient re-training. We also introduce a novel pruning algorithm, compensation-aware pruning, in better collaboration with pruning compensation than existing pruning algorithms. Last, we automate the whole pipeline by structural search with step constraint. Our method can achieve competitive pruning performance on DCNN with significantly improved efficiency.

\section{Related Works}
\label{sec:related_works}

In this section, we briefly review a series of works towards pruning convolutional neural networks and some recent discussions about the effectiveness and efficiency of pruning. We focus on structured pruning of DCNN, i.e. pruning at kernel/channel level. Unstructured pruning(pruning at a parameter-level granularity) results to irregular sparsity of CNN kernel weights, which is incompatible with modern software or hardware implementation of DCNN, hence excluded from our review.

\subsection{Pruning by Heuristics}

An intuitive approach to pruning kernels is to score the importance of kernels by some heuristic-base metrics and remove the least important kernels according to the specified sparsity rate.

Data-free heuristics mainly focus on the magnitude of the uniqueness kernels without consideration on the input for the kernels. For example, Li \textit{et al.}\cite{li2016l1norm} use L1 to evaluate the importance of the kernel. This heuristic is in line with the regularization terms on kernels, which penalize kernels with large norm, and trained kernels with larger magnitude should have greater importance. He \textit{et al.}\cite{he2019filter} introduce a method to pruned redundant kernels by geometric median. Mussay \textit{et al.}\cite{mussay2019coreset} propose to select kernels by the core-set, a small representative subset of all kernels, to preserve the diversity of kernels.

To evaluate the importance of kernels during inference, a number of works incorporate the analysis on features to the evaluation of kernels. Measuring the usage of kernels by the percentage of zero activation, Hu \textit{et al.}\cite{hu2016trimming} propose to evaluate kernel importance of ReLU-activated networks by APoZ(Average Percentage of Zero) of activations. To minimize the oracle loss(a type of reconstruction loss) of layer output after pruning, Molchanov \textit{et al.}\cite{molchanov2016pruning} estimate the loss of a pruning decision by the gradient. Yu \textit{et al.}\cite{yu2018nisp} measure neuron importance by propagating the score between layers. Some recent works in this direction propose different measurement to estimate the information loss of pruning and the corresponding pruning algorithm to minimize the loss, such as synaptic flow\cite{tanaka2020synaptic}, layer-wise OBS\cite{dong2017lobs}, importance estimation\cite{molchanov2019importance}, lookahead\cite{park2020lookahead}, and HRank\cite{lin2020hrank}. 

\subsection{Pruning by Learned Policy}

A more flexible and automated approach towards kernel/channel pruning is to learn a pruning policy from data. The learned policy yield decisions in the pruning step and update the policy by some feedback collected during re-training. The policy can either be an explicit agent rewarded by model-level accuracy and computational cost, or an implicit regularizer that is effective during re-training at neuron level.

He \textit{et al.}\cite{he2018amc} implement a pruning system by reinforcement learning where an agent is trained to prune channels and rewarded by the unification of low training loss and low computational cost. Lin \textit{et al.}\cite{lin2019towards} train a pruning agent(generator) and a discriminator by generative adversarial learning. To another end, Lin \textit{et al.}\cite{lin2020abcpruner} propose a framework of automatic search for layer sparsity rate by bee colony algorithm. Gao \textit{et al.}\cite{gao2021performance} propose to estimate the performance of pruned models by a stand-alone neural network to guide pruning. Tang \textit{et al.}\cite{tang2021manifold} propose to select channels by the manifold relationship between instances and the pruned models. Yeom \textit{et al.} propose a criterion inspired by model interpretability to estimate parameter importance by layer-wise relevance propagation.

As to implicit policy, Liu \textit{et al.}\cite{liu2017slimming} exploit batch normalization layers in DCNN to deactivated kernels with smallest scaling factors. He \textit{et al.}\cite{he2018soft} set soft filter mask instead of commonly used hard masks to gradually weaken the effect of pruned filters. Similarly, Ding \textit{et al.}\cite{ding2020lossless} gradually reduce the dependency on unimportant kernels by resetting the gradients during re-training. Zhuang \textit{et al.}\cite{zhuang2020polarize} polarize the regularization to make more discriminate pruning decisions. Li \textit{et al.}\cite{li2021probablistic} model a posterior distribution of filters pruning decisions to maximize a performance metric.

\subsection{Effectiveness and Efficiency of Pruning}

In company with the development of model pruning, most studies have been considering the channel/kernel selection algorithm as the most significant step until a study by Frankle \textit{et al.}\cite{frankle2018lottery}, which proposes \textit{Lottery Ticket Hypothesis} and presents considerable experimental evidence. A direct conclusion from this work is that re-training a pruned sparse network does not have better performance than training the sparse network from random initialization. This result denies the effectiveness of pruning, because it's the sparsity rate settings instead of the channel/kernel selection that decides the model structure. If the cost of re-training is comparable with training from scratch, it will be more efficient to perform structural search at the beginning than pruning after pre-training. Follow-up works by Liu \textit{et al.}\cite{liu2018rethinking} and Wang \textit{et al.}\cite{wang2021redundancy}, for example, present extensive results showing that the structure(sparsity of layers) is more important than selecting kernels/channels.

Renda \textit{et al.}\cite{renda2019comparing} provide experimental results suggesting that the training strategy of re-training is more significant to the final result. It also put previous pruning methods in doubt, because the re-training settings in most comparisons are not strictly controlled.

We are motivated partly by this recently heated debate over the effectiveness of re-training for accuracy recovery. If the efficiency of accuracy recovery is largely improved, then we can perform a more efficient structural search based on the pre-trained model than evaluating the structure by training from scratch.

Besides, data-efficiency is rarely discussed in the field of pruning. Without specification, the default training data for re-training is same as pre-training. Luo \textit{et al.}\cite{luo2020limited} discuss a re-training strategy enhanced by data augmentation and label refinement to fully exploit the data and achieve better performance in a data-limited condition. But such techniques are not novel or specialized for pruning. Further exploration for data-efficient model pruning will benefit its application on models pre-trained by private or expensive data.

\section{Methodology}
\label{sec:methodology}

\begin{figure}[!t]
    \centering
    \includegraphics[width=2.5in]{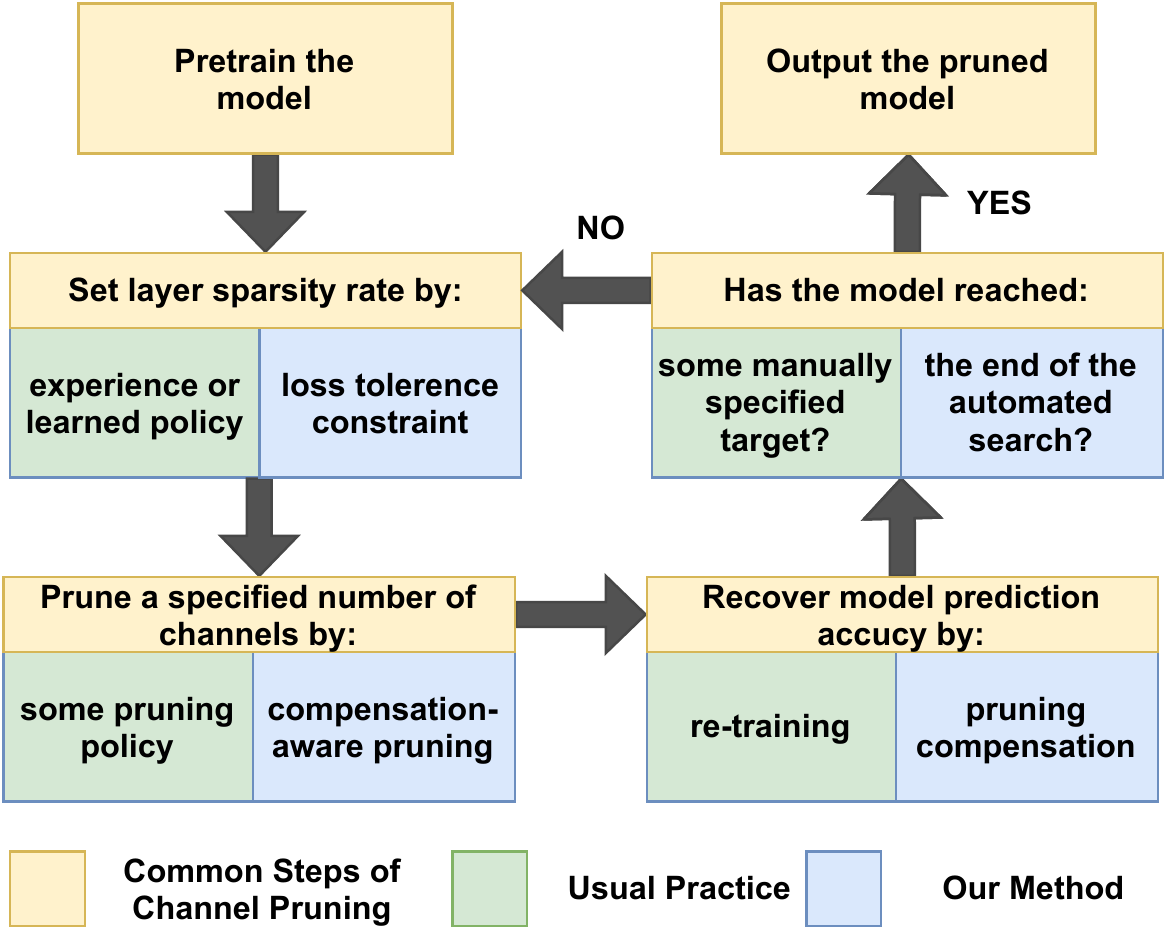}
    \caption{The pipeline of channel pruning(light yellow) implemented by the usual practice(light green) and our method(light blue). The data- and time-inefficient re-training is replaced by pruning compensation, and other steps are also automated by the counter-parts we propose to avoid human interference. }
    \label{fig:pipeline}
\end{figure}

A typical pipeline of channel pruning on convolutional neural networks is illustrated by figure \ref{fig:pipeline}. Apart from the halt condition of the pruning-recovery loop, channel pruning can be decomposed into three core sub-problems:

\begin{enumerate}
    \item How to set a sparsity rate(i.e. the ratio of channels to prune/preserve) for each layer?
    \item How to choose a subset of channels to prune?(channel selection)
    \item How to recover the model prediction accuracy after pruning?
\end{enumerate}

The answers given by most works follow a common paradigm. For sub-problem 1, setting layer sparsity empirically or by learned sparsity posterior(e.g. an additional learned regularizer). For sub-problem 2, prune channels/filters by some heuristics(e.g. magnitude, feature variance, kernel similarity, \textit{etc}) or by a learned channel/filter importance scorer/masker. For sub-problem 3, recovering model prediction accuracy by retraining from the remaining parameters(also called fine-tuning) or random initialization(referred as rewinding in some works).

The most computationally inefficient and data-expensive step of the usual pipeline(green in figure \ref{fig:pipeline}) lies at the recovery of model prediction accuracy(sub-problem 3), which requires sufficient training data, GPU hours and engineering experience for training. To eliminate this cost and meet a low computational budget, we propose to recover model inference by pruning compensation, which is elaborated in subsection \ref{subsec:pruning_compensation}. Pruning compensation avoids iterative parameter optimization and, instead, presents a closed-form solution, to significantly reduce the cost of accuracy recovery. 

We also find that existing heuristic-based pruning methods are not suitable for sub-problem 2 in collaboration with pruning compensation. The recovery capacity of pruning compensation is decided by the information loss and the weight of the pruned channel together. Most existing pruning methods do not reflect the unification of the two properties. Therefore, we propose compensation-aware pruning(CaP), a solution seamlessly coupled with pruning compensation, clarified in subsection \ref{subsec:cap}. CaP searches for an optimal pruning decision using a greedy strategy to minimize the information loss of pruning compensation.

The fast accuracy recovery enabled by pruning compensation significantly speeds up the iteration of model structure search(sparsity of layers). To further automate our pruning workflow, we propose binary structural search with step constraint in subsection \ref{subsec:search} to solve sub-problem 1. Users are freed from tuning hyper-parameters in this process, except setting an explicit tolerance of accuracy drop.

\subsection{Problem Formulation and Notations}

\subsubsection{Channel Pruning Problem of Model}

Generally, a channel pruning problem over neural networks can be formulated as a dual-objective optimization problem as:

\begin{align}
        & \sigma = \{\sigma^{(1)}, \sigma^{(2)}, ... , \sigma^{(L)}\}, \sigma^{(j)}\in[0, 1) \nonumber\\
        & S = \{S^{(1)}, S^{(2)}, ... , S^{(L)}\}, S^{(j)}\subseteq  C^{(j)} \nonumber\\
        & \hat{f} = p(f; \sigma, S) \nonumber\\
        & \min_{\sigma, \hat{\theta}, S} [L(\hat{f}(D;\hat{\theta})), \sum_{j=1}^{L} v^{(j)}(1-\sigma^{(j)}) ] \nonumber\\
        & s.t. \quad |S^{(j)}| \leq (1 - \sigma)|C^{(j)}|, \quad \forall j \in [1, ..., L]
        \label{eq:global_channel_pruning}
\end{align}

The first objective in equation \eqref{eq:global_channel_pruning} is a learning objective on a model $\hat{f}$ parameterized with $\hat{\theta}$, trained on a dataset $D$, and evaluated by a loss function $L$. This learning objective also involves a model selection problem through the pruning operation $p$, which prune each of the $L$ parameterized (convolutional or fully-connected) layers at a sparsity rate $\sigma^{(j)}$ and by a channel selection decision $S^{(j)}$. $S^{(j)}$ are retained channels and should be a subset of $C^{(j)}$(all input channels), while $C^{(j)} - S^{(j)}$ are pruned channels. A very important prior is that the parameters $\theta$ of the un-pruned pre-trained model $f$ has already obtained satisfactorily low loss value, which can be leveraged for learning. 

The second objective in equation \eqref{eq:global_channel_pruning} is a target to reduce the computational cost of model inference. $v^{(j)}$ is the computational workload of the $j$-th layer, normally represented by the number of floating-point operations. It is shrunk proportionally by the layer sparsity rate $\sigma^{(j)}$.

\subsubsection{Intra-layer Channel Pruning Problem}

Common solutions to this dual-objective problem often involves a model-level gradient-based optimization of all the parameters($\sigma$, $\hat{\theta}$, and a "soft" version of $S$). But we propose to solve this problem from an intra-layer angle. An intra-layer channel pruning problem is formulated as an optimization problem to minimize the reconstruction loss of layer output after pruning. In this context, a layer includes a fully-connected(or a convolution layer) together with the following batch normalization operation and non-linear activation function. Then a intra-layer channel pruning problem is:

\begin{align}
    & \min_{\hat{\textbf{W}}, \hat{\textbf{b}}, S} \mathbb{E} ||g(\textbf{Y}) - g(\hat{\textbf{Y}})||^{2} \nonumber \\
    & s.t.\quad |S| \leq (1 - \sigma)|C| \quad and \quad S \subset C \label{eq:intra_layer_channel_pruning}
\end{align}
where $ \textbf{Y} = \textbf{X}_C^{T}\textbf{W} + \textbf{b}$ and $\hat{\textbf{Y}} = \textbf{X}_S^{T}\hat{\textbf{W}} + \hat{\textbf{b}}$. In equation \eqref{eq:intra_layer_channel_pruning}, random variable $\textbf{X}_C \in \mathbb{R} ^ {|C| \times 1}$ is the complete input channels($C$ is an ordered set denoting all indexed input channels), while $\textbf{X}_S \in \mathbb{R} ^ {|S| \times 1}$ is a subset of $\textbf{X}_C$ selected by the pruning algorithm. $\textbf{W}\in\mathbb{R} ^ {|C| \times N}, \textbf{b}\in\mathbb{R}^{1\times N}$ and $\hat{\textbf{W}}\in\mathbb{R} ^ {|C| \times N}, \hat{\textbf{b}}\in\mathbb{R}^{1\times N}$ are layer parameters before and after channel pruning, where $N$ is the number of output channels. $\textbf{Y}\in\mathbb{R}^{1\times N}$ and $\hat{\textbf{Y}}\in\mathbb{R}^{1\times N}$ are output of the layer before and after pruning. A pre-defined layer sparsity rate $\sigma \in [0, 1)$ controls the number of input channels to be pruned. A non-linear function $g$ denotes the following batch normalization operation(if any) and activation function.

For simplicity, we have formulated the global and intra-layer channel pruning problem in a fully-connected layer setting. The problem and our solutions can be generalized to convolutional layers straightforwardly. In practice, $N$ convolutional kernels of shape $\mathbb{R} ^{C_{in}\times k \times k \times N}$ can be vectorized to $\mathbb{R} ^{(k^2C_{in})\times N}$, same are the input features within the sliding window of convolution at each point. Such a "flattened" view of channels and kernels is equivalent to the fully-connected layer discussed above. It's also applicable to generalizing the algorithms in later sections to convolutional layers. 

\subsection{Pruning Compensation}
\label{subsec:pruning_compensation}

\begin{figure}[!t]
    \centering
    \includegraphics[width=2.5in]{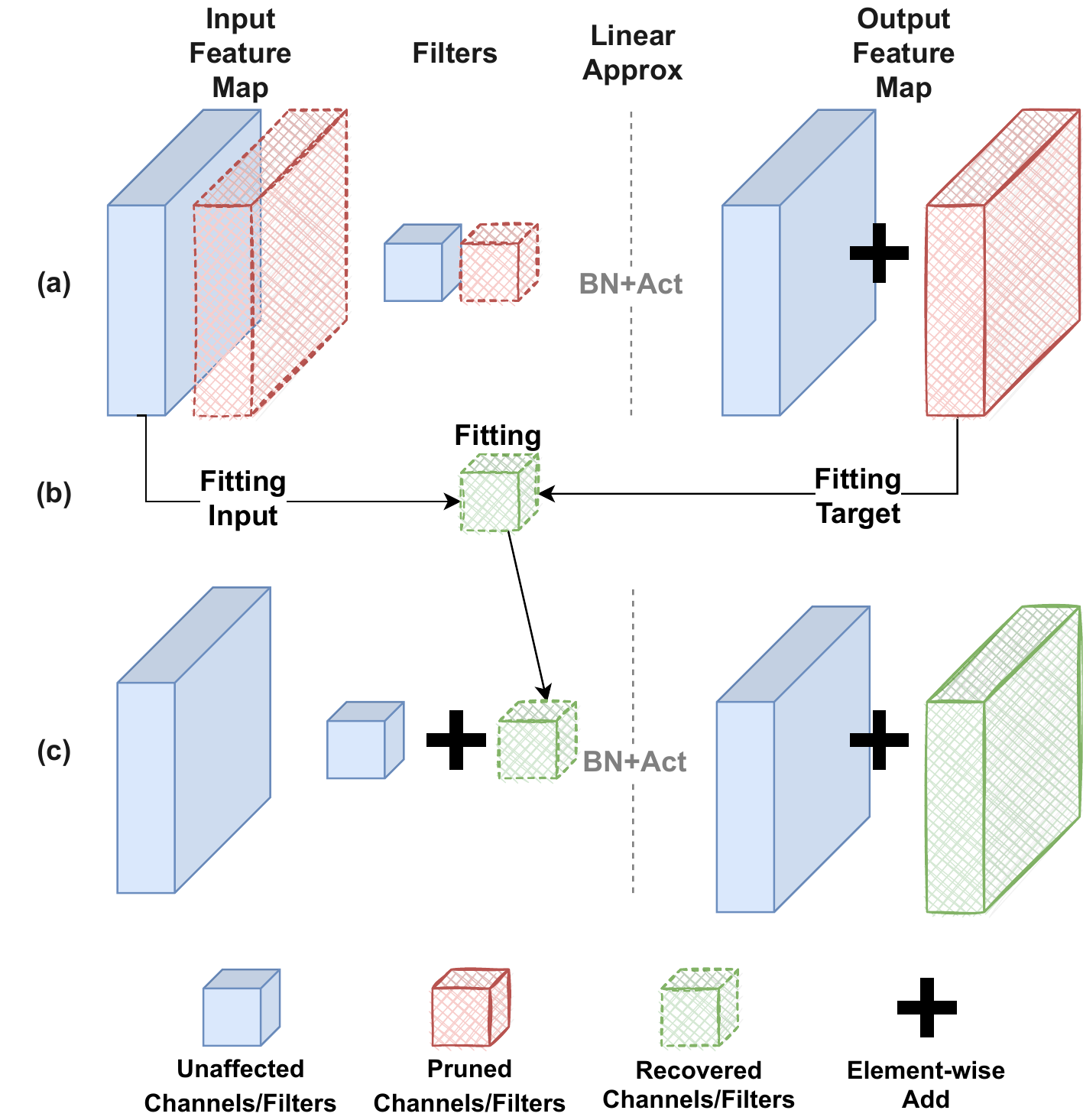}
    \caption{An illustration of pruning compensation. (a) The red channel as well as the matching parameters are pruned. (b) Pruning compensation fits the pruned output channel on the remaining input channel by a new kernel. (c)The sum of the two kernels works as the compensated kernels to recover the pruned channel. }
    \label{fig:compensation}
\end{figure}

\subsubsection{Motivation}

Retraining is often considered to be the standard process to recover the prediction accuracy of the model after pruning. This process trains the remaining parameters in the pruned model to fit the training data again. However, some recent analyses \cite{frankle2018lottery, liu2018rethinking,renda2019comparing} give experimental evidence to show that retraining the model from pre-trained parameters makes no significant difference from retraining from randomly initialized weights. This result denies the effectiveness of pruning algorithms, because any pruning algorithm performed at a specified sparsity rate results in the identical pruned structure(pre-determined by the sparsity rate distribution among layers). Even if retraining from pre-trained parameters is superior in some cases, pruning algorithm's importance over a well-tuned retraining process is still open to doubt.

What we observe from this discussion is how inefficiently we have been exploiting the pre-trained parameters. We argue that, even if pruning will incur some uncertainty to model predictions, the remaining pre-trained parameters still has the potential to recover the original inference. Thus, pruning compensation is introduced to our pruning framework as an efficient substitute of retraining to recover the prediction accuracy of the pre-trained model. Pruning compensation seeks to recover the intermediate features(layer output) of the pruned model. The primary difference from retraining is that pruning compensation focuses on the local structure(usually one fully-connected or convolution layer followed by batch normalization and activation function), while retraining update parameters globally.

\subsubsection{Algorithm Details}

Pruning compensation recovers layer output by fitting pruned input channels by remaining channels, without imposing structural change or computational overhead to the pruned layer. This technique is pruning-algorithm-agnostic, namely applicable to any channel selection result. As a part of the intra-layer channel pruning problem formulated by equation \eqref{eq:intra_layer_channel_pruning}, pruning compensation minimizes the reconstruction loss via $\hat{\textbf{W}}$ and $\hat{\textbf{b}}$, positing that $S$ is fixed.

In this paper, we render this optimization problem computationally inexpensive. The non-linear function $g$ in the optimization problem in equation \eqref{eq:intra_layer_channel_pruning}(with $S$ fixed) is the most intractable part, so we convert the objective to its linear approximation by Taylor expansion, with the expectation becoming an estimation from data:

\begin{align}
    & \min_{\hat{\textbf{W}}, \hat{\textbf{b}}} \frac{1}{M} \sum_{i=1}^{M} ||g'(\textbf{Y}_i)(\textbf{Y}_i - \hat{\textbf{Y}}_i)||^2 \nonumber\\
    = & \frac{1}{M} \sum_{i=1}^{M} ||g'(\textbf{Y}_i)(\textbf{X}_i^T\textbf{W} + \textbf{b} - \textbf{X}_{S,i}^T\hat{\textbf{W}} - \hat{\textbf{b}})||^2
    \label{eq:intra_layer_channel_pruning_estimation}
\end{align}
where $M$ instances are sampled from the training data for estimation. Such a constraint-free quadratic programming problem has a closed-form solution:

\begin{align}
    & \hat{\textbf{W}} = \mathbf{\Sigma}_{S,S}^{-1}\mathbf{\Sigma}_{S,C}\textbf{W} \nonumber \\
    & \hat{\textbf{b}} = \mathbf{\mu}_C^T \textbf{W} + \textbf{b} - \mathbf{\mu}_S^T \hat{\textbf{W}} \label{eq:intra_layer_channel_pruning_solution}
\end{align}

The estimation of the feature expectation $\mathbf{\mu}_C \in \mathbb{R}^{|C|\times1}$, $\mathbf{\mu}_S \in \mathbb{R}^{|S|\times1}$ and covariance $\mathbf{\Sigma}_{S,S}\in\mathbb{R}^{|S|\times|S|}$, $\mathbf{\Sigma}_{S,C}\in\mathbb{R}^{|S|\times|C|}$ are computed by:

\begin{align}
    & \mathbf{\mu}_C = \frac{1}{M}\sum_{i=1}^{M}g'(\textbf{Y}_i)^{2}\textbf{X}_{C,i}  \quad \mathbf{\mu}_S = \frac{1}{M}\sum_{i=1}^{M}g'(\textbf{Y}_i)^{2}\textbf{X}_{S,i} \nonumber\\
    & \mathbf{\Sigma}_{S,S} = \frac{1}{M-1}\sum_{i=1}^{M}g'(\textbf{Y}_i)^{2}\textbf{X}_{S,i}\textbf{X}_{S,i}^T - \mathbf{\mu}_S\mathbf{\mu}_S^T \nonumber\\
    & \mathbf{\Sigma}_{S, C} = \frac{1}{M-1}\sum_{i=1}^{M}g'(\textbf{Y}_i)^{2}\textbf{X}_{S,i}\textbf{X}_{C,i}^T - \mathbf{\mu}_S\mathbf{\mu}_C^T
    \label{eq:statistics_estimation}
\end{align}

In practice, pruning compensation for each layer consists of three steps:

\begin{itemize}
    \item Collect the layer inputs and outputs by forward-propagation on a sampled subset of the training data $D$. \\
    \item Estimated of the required statistics by equation \eqref{eq:intra_layer_channel_pruning_estimation}. \\
    \item Replace the original parameters with the compensated parameters $\hat{\textbf{W}}, \hat{\textbf{b}}$ computed via equation \eqref{eq:intra_layer_channel_pruning_solution}. \\
\end{itemize}

Many CNN architectures have bias-free convolutional or fully-connected layers. But a bias term is necessary for pruning compensation to adjust the mean of the post-compensation layer output $\hat{\textbf{Y}}$. So we append a bias term to bias-free compensated layers, which slightly change the model structure. The computational overhead incurred by an additional bias term is insignificant compared to the intensive convolution operation.

\subsubsection{Discussions}

We find that the introduction of $g'(\textbf{Y}_i)$ is critical to the recovery power of pruning compensation. In approximation to the batch-normalization and non-linear function($g$) that shift, rescale and warp $\textbf{Y}$, $g'(\textbf{Y}_i)$ weighted each instance $\textbf{Y}_i$ differently in the reconstruction loss. For example, to compensate a sigmoid-activated convolutional layer, $g'(\textbf{Y}_i)$'s values are close to $0$ for those with large absolute values. As a result, instance $\textbf{Y}_i$ has little weight in the reconstruction loss as equation \eqref{eq:intra_layer_channel_pruning_estimation}. From a post-hoc perspective, this little weight reflects the fact that $g(\textbf{Y}_i)$ has minor change after pruning when $\textbf{Y}_i$ has large absolute values. In contrast, instances with large $|g'(\textbf{Y}_i)|$ are hard samples for optimization of the reconstruction loss, being heavily weighted.

Pruning compensation is of better time and data efficiency than retraining. The major time-consuming step of pruning compensation is the collection of layer input/output over the training data, which requires model forward-propagation on training data for one pass. This cost is significantly lower than retraining, which computes forward/backward-propagation and update the weights for tens, hundreds or even thousands of epochs.

What's more, the estimation doesn't necessarily need the whole training dataset. A reasonably small portion of training data can still produce a stable estimation. When the training data is inaccessible, the estimation can even be made on another more accessible data source of similar distribution, due to the transferability of DCNN features. For example, when we prune a model pre-trained on ImageNet but we cannot use the original training data, we can use PASCAL VOC images as an alternative data source for estimation. It's a useful quality on tackling models pre-trained on private or scarcely accessible data.

\subsection{Compensation-aware Pruning}
\label{subsec:cap}

\subsubsection{Renewed Problem Formulation}

Pruning compensation provides a closed-form solution to the intra-layer channel pruning problem on variables $\hat{\textbf{W}}$ and $\hat{\textbf{b}}$, as equation \eqref{eq:intra_layer_channel_pruning}, under any channel selection decision $S$. Thus, we can solve for $S$ by replacing the determined variables $\hat{\textbf{W}}$ and $\hat{\textbf{b}}$ with their closed-form solution(solutions in equation \eqref{eq:intra_layer_channel_pruning_solution} substituting variables in equation \eqref{eq:intra_layer_channel_pruning}), and the problem is reduced to a combinatorial optimization problem in equation \eqref{eq:CaP}, where $\textbf{W} = [\textbf{w}_1, \textbf{w}_2, ..., \textbf{w}_N]$ are weights before compensation:

\begin{align}
    \min_{S} & \sum_{k=1}^{N} \textbf{w}_k^T\mathbf{\Sigma}_{C,C}\textbf{w}_k - \textbf{w}_k^T\mathbf{\Sigma}_{C,S}\mathbf{\Sigma}_{S,S}^{-1}\mathbf{\Sigma}_{S,C}\textbf{w}_k \nonumber\\
    s.t.& \quad|S| \leq (1 - \sigma)|C| \quad and \quad S \subset C
    \label{eq:CaP}
\end{align}

As this pruning method presumes that the lost information in pruned channels could be compensated afterwards, we name this method as compensation-aware pruning(CaP). CaP provides a unified approach to select channels by information loss and channel weight together.

To better interpret the implicit channel selection rule under equation \eqref{eq:CaP}, we explain via two extreme cases. In the two extreme cases, we assume that the diagonal entries of $\mathbf{\Sigma}_{C,C}$ are all $1$, i.e. every channel's variance equals $1$.

In the first case, we assume that the input channels are independent from each other, then the covariance $\mathbf{\Sigma}_{C,C}$ will be an identity matrix. In this case, the importance of the $i$-th channel will be determined by the corresponding weight $\textbf{w}_{k, i}$ only. Larger the $|\textbf{w}_{k, i}|$, higher the importance. This non-trivial situation is equivalent to magnitude-base pruning.

In the second case, we assume that each channel is equally weighted, i.e. each entry of $\textbf{W}$ is $1$, and channels in $S$ or $C-S$ are independent from each other, i.e. $\mathbf{\Sigma}_{S,S}$ and $\mathbf{\Sigma}_{C-S,C-S}$ are identity matrix. Note that $\mathbf{\Sigma}_{C-S, S}$ is not necessarily zero matrix. The reconstruction loss will then be determined by the sum of $||\mathbf{\Sigma}_{C-S, S}||^2_2$. Higher the absolute values in $\mathbf{\Sigma}_{C-S, S}$, more replaceable are channels in $C-S$; less information lost by $C-S$, less the reconstruction loss.

These two cases interpret respectively how CaP select channels by weights or covariance. In a non-trivial setting, the channel selection decision $S$ is made in unification of both two properties.

\subsubsection{Greedy Search}

CaP as a combinatorial optimization problem is NP-hard, and the evaluation of an arbitrary solution $S$ requires an inversion of an $|S|\times|S|$ matrix, which is expensive. For example, a layer with $N=1024$ input channels pruned at sparsity rate $\sigma=0.5$ will require $\binom{1024}{512}$ times of $512 \times 512$ matrix inversion if performing exhaustive search. We propose a greedy algorithm to find a sub-optimal solution to this problem.

Compensation-aware pruning is detailed in algorithm \ref{alg:cap}. We iteratively expand the solution $S$ by incorporating the best of the remaining channel to minimize a reconstruction loss conditioned on current $S$. The procedure continues util the sparsity constraint becomes effective. The computational complexity of our algorithm is reduced to $\mathcal{O}(N^5)$(the greedy search from line 4 to 7 in algorithm \ref{alg:cap} has quadratic complexity, and each evaluation of the reconstruction loss in line 5 has cubic complexity to compute $\mathbf{\Sigma}_{S,S}^{-1}$).

\begin{algorithm}[!t]
    \caption{Compensation-aware Pruning}
    \label{alg:cap}
    \begin{algorithmic}[1]
        \State \textbf{Input:} sparsity rate($\sigma$), input channels($C$)
        \State \textbf{Output:} selected retained channels($S$)
            \State $S \leftarrow \emptyset$
            \While{$|S| < (1-\sigma)|C|$}
                \State Find $i\in C-S$ that minimize $loss(S\cup\{i\})$ \Comment loss defined in eq \eqref{eq:CaP}
                \State $S \leftarrow S \cup \{i\}$
            \EndWhile
    \end{algorithmic}
\end{algorithm}

We further accelerate CaP by exploiting the computational shortcut of the inversion of an augmented matrix. For a temporary channel selection $S$, adding a channel $c$ that makes $\hat{S}=S\cup\{c\}$ will involve the computation of $\mathbf{\Sigma}_{\hat{S}, \hat{S}}^{-1}$. We find an efficient way to perform this inversion by exploiting the known $\mathbf{\Sigma}_{S, S}^{-1}$ and prove it in Appendix \ref{sec:appendix_a}. The computational complexity is reduced to $\mathcal{O}(N^4)$.

Pruning compensation and CaP both rely on $\mathbf{\Sigma}_{S, S}^{-1}$. We need to ensure the selected channels $S$ lead to a non-singular $\mathbf{\Sigma}_{S, S}$. Because the feature maps $X$ are always transformed by a non-linear activation function, $\mathbf{\Sigma}_{S, S}$ is not singular in most of the time. To tackle exceptions, we exclude any channel with near-zero variance(diagonal entries in $\mathbf{\Sigma}_{C, C}$) from $S$. Also, in CaP(algorithm \ref{alg:cap}), a channel $i$ will not be incorporated into $S$ if the inversion of $\mathbf{\Sigma}_{\hat{S}, \hat{S}}$ fails because of its singularity or a non-positive eigenvalue arises. This rule guarantees that $\mathbf{\Sigma}_{S, S}$ is always positive definite, which facilitates the acceleration of CaP in Appendix \ref{sec:appendix_a} using Cholesky factorization.

\subsection{Binary Structural Search with Step Constraint}

Having found a solution to intra-layer channel pruning problem in equation \eqref{eq:intra_layer_channel_pruning}, we focus on the last variable, layer-wise sparsity rate $\sigma$, in the global(model-level) channel pruning problem in equation \eqref{eq:global_channel_pruning}. As the accuracy recovery step is dramatically accelerated by pruning compensation, fine-grain structural search can be performed more efficiently.

\begin{algorithm}[!t]
    \caption{Binary Structural Search with Step Constraint}
    \label{alg:search}
    \begin{algorithmic}[1]
        \State \textbf{Input:} model(directed acyclic computational graph) layers($M$), number of layers to prune($L$), accuracy drop tolerance($\tau$), search steps($k$).
        \State \textbf{Output:} $M$
        \State \textbf{Initialization:} $\sigma^{(i)}\leftarrow 0, i\in\{0, 1, ..., L-1\}$
        \For{$i \leftarrow 0$; $i < L$; $i++$}
            \State $\tau_{i} \leftarrow \tau * (i+1)/L$
            \State $\sigma_{min} \leftarrow 0$
            \State $\sigma_{max} \leftarrow 1$
            \For {$j\leftarrow 0$; $j < k$; $j++$}
                \State $\sigma \leftarrow (\sigma_{min} + \sigma_{max})/2$
                \State $M[i] \leftarrow CaP(\sigma^{(i)}, M[i])$ \Comment algorithm \ref{alg:cap}
                \State $M[i] \leftarrow Compensate(M[i])$
                \State $accDrop \leftarrow TestAccuracy(M)$
                \If{$accDrop \geq \tau$}
                    \State $\sigma_{max} \leftarrow \sigma$
                \Else
                    \State $\sigma_{min} \leftarrow \sigma$
                \EndIf
            \EndFor
            \State $\sigma^{(i)} = \sigma_{min}$
        \EndFor
    \end{algorithmic}
\end{algorithm}

Global channel pruning(equation \eqref{eq:global_channel_pruning}) is a dual-objective optimization problem, which can be solved by setting a hard constraint on one objective and optimize the other one. In our method, we allow the user to set a constraint on the loss of the pruned model. Structural search is performed to minimize the computational cost of the model while retaining the loss below the user-set constraint. To make the interface more user-friendly, we use model prediction accuracy(e.g. classification precision or recall) on the validation dataset instead of some commonly used loss function, such as cross-entropy loss or mean-square error. The user will need to set a tolerance value to reject sparse models whose prediction accuracy on the validation set drops too heavily due to over-pruning. Algorithm \ref{alg:search} provides a full procedure of binary structural search with step constraint.

\begin{figure}[!t]
    \centering
    \includegraphics[width=2.5in]{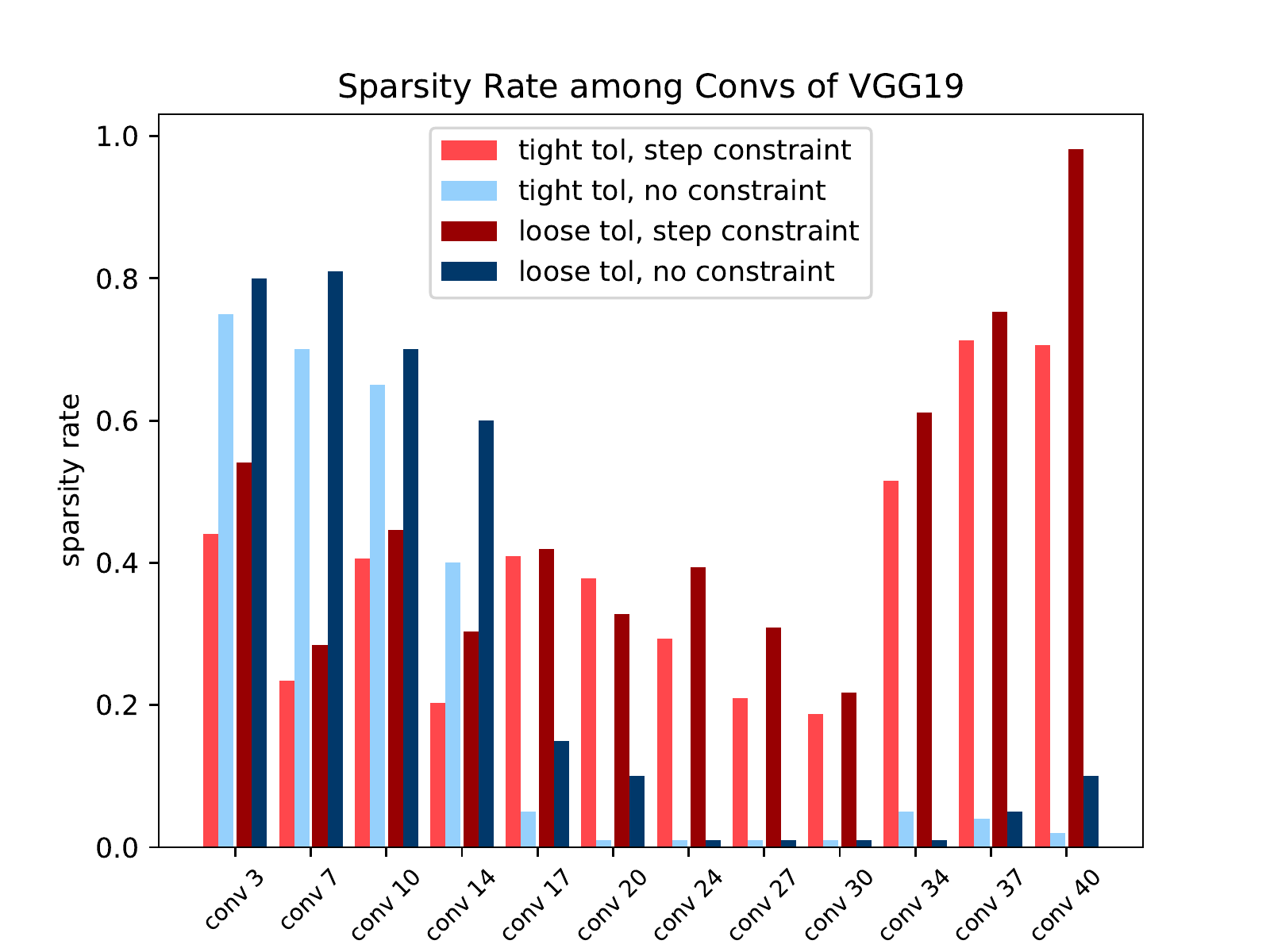}
    \caption{Structural search results on VGG-19 with step constraint or global constraint(user-set tolerance). \textsl{tight/loose tol} are respectively 1\% and 3\% tolerance of accuracy drop. Without a step constraint, the search process is prone to over-pruning at the early stage and a sub-optimal structure.}
    \label{fig:sparsity}
\end{figure}

We also introduce a step constraint to control the search path. Without a step constraint, the first-visited layers tend to be over-pruned(too high sparsity rate), leaving scarce room for pruning later layers. The trade-off among the sparsity of all layers will be imbalanced, giving too much priority to early pruned layers. Figure \ref{fig:sparsity} illustrates an example on VGG-19. So we put a step constraint into the search process. A step constraint is the split of the user-set tolerance into each layer. Without prior knowledge, we divide the tolerance uniformly among layers. This step constraint prevents over-pruning at the early stage.

Our method does not specify a visiting order of layers. Theoretically, the visiting order has some uncertain impact on pruning. Note that the mean vector and the covariance matrix used for pruning compensation is estimated for only one time. When a layer is pruned, it will always cause some perturbation on the estimation of dependent layers, unlikely to be foreseen. For two layers $L^{(1)}$ and $L^{(2)}$, where $L^{(2)}$'s input is dependent on the output of $L^{(1)}$: if $L^{(2)}$ is pruned before $L^{(1)}$, the later pruning on $L^{(1)}$ will bring about more uncertainty to the input of $L^{(2)}$, possibly significant enough to overturn the made pruning decision; if $L^{(1)}$ is pruned before $L^{(2)}$, when $L^{(2)}$ is pruned and compensated, the mean and variance are estimated from the output of un-pruned $L^{(1)}$, which is not in line with the current pruned distribution. Our experiment reveals that specifying a visiting order is not very helpful. Thus, without specification, we use a bottom-up order(the visiting order of model inference) by default in our experiments.

\section{Experiments}
\label{sec:experiments}

\subsection{Experiment Settings}

\subsubsection{Datasets}

We evaluate our method and other pruning algorithms on CIFAR-10, CIFAR-100\cite{krizhevsky2009cifar}, and ILSVRC-2012(ImageNet)\cite{russakovsky2015imagenet}. Although channel pruning is applicable to DCNN trained for any visual task(e.g. classification, detection, segmentation, tracking, synthesis), we choose these three single-label image classification tasks to evaluate pruning methods by convention. Image classification, especially large-scale ones like ImageNet classification, is usually considered to be the baseline task among all visual tasks, and most other tasks use backbone models originally designed for image classification. The choice of benchmarks ensures the generalizability of the pruning method to other tasks.

CIFAR-10/CIFAR-100 have 10 and 100 classes respectively, both consist of $50,000$ training images and $10,000$ testing images. The 3-channel RGB images have $32\times32$ resolution, and the label distribution is equally balanced. We use $40,000$ randomly sampled training images for pre-training and the estimation of feature statistics for pruning compensation, and the other 10,000 training set images form a validation set to guide the structural search(to meet the step constraint of accuracy drop). The reported classification accuracy in the following experiments is calculated from the predictions on the test set, which is not involved in pre-training or pruning.

ImageNet is a large scale benchmark compared to CIFAR-10/CIFAR-100, consisting of a train-split of $1.28$ million images and a val-split of $50,000$ images. The $1,000$ classes have balanced label distribution in both splits. The images have various resolutions, so the images are resized to $256\times256$ for batched training and testing. We pre-train the models on 95\% of the train-split(randomly sampled, referred as the training set in this section), and perform validation on the remaining 5\% train-split. The final classification accuracy reported is based on the val-split.

Besides, our experiments make use of the training images($11,540$ images) in PASCAL VOC 2012\cite{everingham2010pascal}, to validate that the estimation of the feature statistics can rely on another data source. The images in PASCAL VOC 2012 are similar natural images as ImageNet, both mainly collected from the Internet and manually annotated, except that the 20 object classes of PASCAL VOC are different from the $1,000$ classes ImageNet and there is not always a dominant object at the center as ImageNet images. We resize PASCAL VOC 2012 images to $256\times256$ as well.

\subsubsection{Baseline Models}

We evaluate our method and other pruning algorithms on VGG-16/19\cite{simonyan15vgg} and ResNet-50\cite{he2016resnet}, not only because they are the most common testbeds for channel pruning, but also for their extensive use in most visual tasks.

VGG architecture features sequentially connected convolutional layers, followed by 3 fully connected layers for classification. Each convolutional or fully connected layer is rescaled by a batch normalization\cite{ioffe2015batchnorm} layer and activated by a ReLU function\cite{nair2010rectified}. VGG-16 and VGG-19 have 13 and 16 convolutional layers of $3\times3$ kernels, and the channel number range from 64 to 512.

ResNet has more complex layer connection than VGG. A residual block has two branches relying on one feature map. This is undesirable for channel pruning, because a channel can be removed only when it's pruned in all dependent layers. ResNet-50 has 53 convolutional layers(each has 64 to 2048 channels) and only one fully-connected layer.

In practice, we do not prune the channels the first convolutional layer, which only has 3 input channels. Also, we do not prune the fully-connected layers because they are closely connected to the model output but have inconspicuous contribution to the total computational cost.

\subsubsection{Compared Methods}

Our method is evaluated in comparison with a number of state-of-the-art pruning methods, including Thi-Net\cite{luo2018thinet}, Generative Adversarial Learning(GAL)\cite{lin2019towards}, Meta-Pruning\cite{liu2019metapruning}, FPGM\cite{he2019filter}, Importance Estimation(IE)\cite{molchanov2019importance}, HRank\cite{lin2020hrank}, Discrimination-aware Channel Pruning\cite{liu2021discrimination}. The cited pruning results of the compared methods are reported in the original papers. We implement HRank, FPGM, IE for further analysis, referring the code officially released by the authors.

\subsubsection{Performance}

The performance of the pruned models is evaluated via prediction accuracy on the test dataset and the number of floating-point operations(FLOPs).Top-1 and top-5 classification accuracy on CIFAR-100 and ImageNet are reported, while only top-1 accuracy is reported on CIFAR-10(10-class). The accuracy is the average classification precision of all classes. Because the labels are balanced in all benchmarks, the class precisions are not weighted. FLOPs measures the computational cost of model inference by the amount of 32-bit floating-point operations conducted by GPU. It's positively correlated to the inference time of the model. The total FLOPs is the sum of the layer FLOPs. For a layer whose input feature map is a $C_{in}\times W\times H$ tensor: if the layer is a convolutional layer with output channels $C_{out}$, kernel size $k$, and stride $s$, the layer FLOPs is $(k^2WHC_{in}C_{out})/s^2$; if the layer is a batch-normalization layer, the layer FLOPs is $2WHC_{in}$; if the layer is a non-linear activation function, whose FLOPs on a single input is $n$, the layer FLOPs is $nWHC_{in}$.

We report the GPU hours and memory consumption of re-training and pruning compensation to demonstrate the efficiency of our method in subsection \ref{subsec:compensation}. Our experiments are conducted on multiple TITAN RTX GPUs with 24GB memory. The reported GPU hours is the sum of computational time on all used GPUs. We fully exploit the memory if possible to maximize the processing speed of both re-training and pruning compensation.

\subsection{Comparison with Other Methods}

\begin{table}[!t]
    \setlength{\tabcolsep}{1pt} 
    \renewcommand{\arraystretch}{1.3}
    \caption{Pruning results on ResNet-50 pre-trained on ImageNet. This table reports the top-1/top-5 classification accuracy, accuracy drop and FLOPs drop of the pruned model. Each method has different baseline accuracy and same baseline FLOPs. The processing time is estimated by our hardware environment and the reported pruning procedure.}
    \label{tbl:prune_imagenet}
    \centering
    \begin{tabular}{c || c | c | c | c }
    \hline
    \footnotesize{\textbf{Method}} & \footnotesize{\textbf{Top1-acc(\%)}} & \footnotesize{\textbf{Top5-acc(\%)}} & \footnotesize{\textbf{FLOPs drop(\%)}} & \footnotesize{\textbf{Proc Time(h)}}\\
    \hline

    FPGM-mix \cite{he2019filter} & 75.50($\downarrow$0.65) & 92.66($\downarrow$0.21) & 42.20 & 42.67 \\
    FPGM-only \cite{he2019filter} & 75.59($\downarrow$0.56) & 92.63($\downarrow$0.24) & 42.20 & 42.67 \\

    IE \cite{molchanov2019importance} & 74.50($\downarrow$1.68) & N/A & 44.99 & 17.78 \\

    HRank\cite{lin2020hrank} & 74.98($\downarrow$1.17) & 92.33($\downarrow$0.54) & 43.77 & 341.33 \\
    HRank\cite{lin2020hrank} & 71.98($\downarrow$4.17) & 91.01($\downarrow$1.86) & 62.10 & 341.33 \\
    HRank\cite{lin2020hrank} & 69.10($\downarrow$7.05) & 89.58($\downarrow$3.29) & 76.04 & 341.33 \\

    GAL-0.5\cite{lin2019towards} & 71.95($\downarrow$4.20) & 91.94($\downarrow$0.93) & 43.03 & 21.33 \\
    GAL-1\cite{lin2019towards} & 69.88($\downarrow$6.27) & 89.75($\downarrow$3.12) & 61.37 & 21.33 \\
    GAL-0.5-joint\cite{lin2019towards} & 71.80($\downarrow$4.35) & 90.82($\downarrow$2.05) & 55.01 & 21.33 \\
    GAL-1-joint\cite{lin2019towards} & 69.31($\downarrow$6.84) & 89.12($\downarrow$3.75) & 72.86 & 21.33 \\

    MP-0.85\cite{liu2019metapruning} & 76.20($\downarrow$0.40) &  N/A & 26.82 & 22.76 \\
    MP-0.75\cite{liu2019metapruning} & 75.40($\downarrow$1.20) & N/A & 51.22 & 22.76 \\

    ThiNet-70\cite{luo2018thinet} & 74.03($\downarrow$1.27) & 92.11($\downarrow$0.09) & 41.97 & 23.45 \\
    ThiNet-50\cite{luo2018thinet} & 72.03($\downarrow$3.27) & 90.99($\downarrow$1.21) & 55.83 & 23.46 \\
    ThiNet-30\cite{luo2018thinet} & 68.17($\downarrow$7.13) & 88.86($\downarrow$3.34) & 71.50 & 23.46 \\

    DCP\cite{liu2021discrimination} & 74.99($\downarrow$1.02) & 92.20($\downarrow$0.73) & 51.55 & 85.33 \\

    \hdashline
    Ours(tol=0.5\%) & 75.56($\downarrow$0.44) & 92.73($\downarrow$0.13) & 41.39 & 2.51 \\
    Ours(tol=1\%) & 75.11($\downarrow$0.89) & 92.47($\downarrow$0.39) & 46.84 & 2.51 \\
    Ours(tol=2\%) & 74.16($\downarrow$1.84) & 92.08($\downarrow$0.78) & 52.10 & 2.51 \\
    Ours(tol=5\%) & 71.13($\downarrow$4.87) & 90.96($\downarrow$1.90) & 62.58 & 2.51 \\
    \hline
    \end{tabular}\\
\end{table}

\begin{table}[!t]
    \renewcommand{\arraystretch}{1.3}
    \caption{Pruning results on VGG-16 rre-trained on CIFAR-10. This table reports the top1 classification accuracy and FLOPs of the pruned model and its drop from the un-pruned baseline. Each method has different baseline accuracy and same baseline FLOPs. The processing time is estimated by our hardware environment and the reported pruning procedure.}
    \label{tbl:prune_cifar10}
    \centering
    \begin{tabular}{c | c | c | c }
    \hline
    \footnotesize{\textbf{Method}} & 
	\footnotesize{\textbf{Top1-acc(\%)}} & 
	\footnotesize{\textbf{FLOPs drop(\%)}} &
    \footnotesize{\textbf{Proc Time(h)}}\\

    \hline
    HRank\cite{lin2020hrank} & 93.43($\downarrow$0.53) & 53.5 & 2.66 \\
    HRank\cite{lin2020hrank} & 91.23($\downarrow$2.73) & 76.50 & 2.66 \\

    GAL-0.05\cite{lin2019towards} & 93.77($\downarrow$0.19) & 39.60 & 1.11 \\
    GAL-0.1\cite{lin2019towards} & 93.42($\downarrow$0.54) & 45.20 & 1.11 \\

    DCP\cite{liu2021discrimination} & 94.29($\downarrow$0.31) & 50.08 & 2.78 \\

    CAC\cite{chen2020dynamical} & 92.90($\uparrow$0.43) & 51.02 & 1.89 \\

    \hdashline
    Ours(tol=0.5\%) & 93.51($\downarrow$0.45) & 43.23 & 0.20 \\
    Ours(tol=1.0\%) & 93.14($\downarrow$0.80) & 50.90 & 0.20 \\
    Ours(tol=1.5\%) & 92.66($\downarrow$1.30) & 56.77 & 0.20 \\
    Ours(tol=2.0\%) & 92.24($\downarrow$1.72) & 59.72 & 0.20 \\
    Ours(tol=2.5\%) & 91.87($\downarrow$2.09) & 65.61 & 0.20 \\
    \hline
    \end{tabular}
\end{table}

Our method(combining pruning compensation, CaP, and structural search) is compared with other state-of-the-art pruning methods on two benchmarks: ResNet-50 pre-trained and pruned on ILSVRC-2012(ImageNet); VGG-16 pre-trained and pruned on CIFAR-10. Table \ref{tbl:prune_imagenet} and table \ref{tbl:prune_cifar10} report the results. Besides top1/top5 accuracy and FLOPs, we also compare the processing time(represented by GPU hours) needed by each method. The processing time is recorded by running the pruning process reported in the original papers in our hardware environment(mainly decided by the power and the number of GPU). For example, \cite{he2019filter} re-trains pruned ResNet-50 for 60 epochs on ImageNet's training set, so the processing time reported in table \ref{tbl:prune_imagenet} is estimated by running the re-training scheme on our device. In both two experiments, we set different tolerance on top1 accuracy drop to compress the model to different extend. 

The experimental results show that our method has competitive performance among the state-of-the-art pruning methods. In table \ref{tbl:prune_imagenet}, our method reduce 46.84\% of FLOPs in ResNet-50 at the loss of 0.89\% drop on top1 accuracy and 0.39\% drop on top5 accuracy. At the similar degree of compression, IE and HRank cause 1.68\% or 1.17\% drop in top1 accuracy. When a large tolerance(5\%) is specified, the reduction of FLOPs grows to 62.58\%. Our method has comparable results with HRank, who achieves a 4.17\% top1 accuracy drop and 62.10\% FLOPs drop by re-training. In table \ref{tbl:prune_cifar10}, we find that most of the pruning results, either by our method or other method, achieve a FLOPs drop around 50\% at the loss of top1 accuracy around 1\%. CIFAR-10 is a much easier classification task compared to ImageNet, so other re-training based pruning methods can promisingly re-gain a high accuracy by the pruned model. It's difficult to clarify whether the pruning algorithm or the re-training is more responsible for the recovered accuracy, because it's not always rigorously controlled. As our method is more flexible on the choice of tolerance, we report the pruning results with accuracy drop below 2.5\%.

Our method has distinct superiority on efficiency over other methods. All re-training based pruning methods demand tens or even hundreds of GPU hours to recover the accuracy when pruning ResNet-50 pre-trained on ImageNet, while our method needs only 2.51 GPU hours. In table \ref{tbl:prune_cifar10}, the inefficiency of re-training is not very obvious on CIFAR-10, which is a small-scale and low-resolution dataset compared with ImageNet. But our method can still achieve $6\times$ speed-up and reduce the processing time to 0.2 GPU hour. We decompose the total processing time of our method in subsection \ref{subsec:compensation} for more in-depth analysis on the efficiency of our method.

\subsection{Evaluating the Efficiency of Pruning Compensation}
\label{subsec:compensation}

\begin{table}[!t]
    \setlength{\tabcolsep}{1pt} 
    \renewcommand{\arraystretch}{1.3}
    \caption{Comparing pruning compensation and re-training under limited data access. Reporting pruning results of VGG-19 on ImageNet. Top1 accuracy drop tolerance is set to 2\%.}
    \label{tbl:prune_var_stat}
    \centering
    \begin{tabular}{c || c | c | c}
    \hline
    \footnotesize{\textbf{Data Source}} & 
	\footnotesize{\textbf{Top1-acc(\%)}} & 
	\footnotesize{\textbf{Top5-acc(\%)}} & 
	\footnotesize{\textbf{FLOPs drop(\%)}} \\

    \hline
    50\% ImageNet & 72.20($\downarrow$2.02) & 90.75($\downarrow$1.09) & 49.50 \\

    20\% ImageNet & 72.08($\downarrow$2.14) & 90.61($\downarrow$1.23) & 48.48 \\

    10\% ImageNet& 72.17($\downarrow$2.05) & 90.68($\downarrow$1.16) & 48.87 \\

    5\% ImageNet&72.21($\downarrow$2.01) & 90.77($\downarrow$1.07) & 48.69 \\

    100\% VOC & 72.15($\downarrow$2.07) & 90.64($\downarrow$1.20) & 47.93 \\

    50\% VOC & 72.17($\downarrow$2.05) & 90.70($\downarrow$1.14) & 47.94 \\

    10\% VOC & 72.28($\downarrow$1.94) & 90.72($\downarrow$1.12) & 47.59 \\

    Data-free(amount 50\% ImageNet) & 72.50($\downarrow$1.72) & 90.95($\downarrow$0.89) & 24.44 \\
    Data-free(amount 20\% ImageNet) & 72.64($\downarrow$1.58) & 91.04($\downarrow$0.80) & 24.00 \\
    Data-free(amount 10\% ImageNet) & 72.53($\downarrow$1.69) & 90.93($\downarrow$0.91) & 24.44 \\
    Data-free(amount 5\% ImageNet) & 72.48($\downarrow$1.74) & 90.92($\downarrow$0.92) & 20.78 \\

    \hline
    \end{tabular}
\end{table}

\begin{table}[!t]
    \setlength{\tabcolsep}{6pt} 
    \renewcommand{\arraystretch}{1.3}
    \caption{Comparison of the processing time(GPU hours) needed for retraining and our method. Re-training and estimation is performed on the training data. Accuracy drop(Acc-drop) evaluation is performed on the validation data.}
    \label{tbl:runtime}
    \centering
    \begin{tabular}{c || c}
        \hline
        \textbf{Model} & ResNet-50 \\
        \textbf{Training Data} & 1.5m images from ImageNet \\
        \textbf{Validation Data} & 50k images from ImageNet \\
        \hline
        \textbf{Operation} & \textbf{Proc Time(h)} \\
        \hline
        \textbf{Re-training for 100 epochs} & $0.717 \times 100 = 71.71$ \\
        \hdashline
        \textbf{Pruning Compensation} & $1.616 \times 1 = 1.616$ \\
        \textbf{CaP} & $0.208 \times 1 = 0.208$ \\
        \textbf{Acc-drop evaluation} & $0.014 \times 48 = 0.672$ \\

        \hline
        \hline

        \textbf{Model} & VGG-16 \\
        \textbf{Training Data} & 40k images from CIFAR-10 \\
        \textbf{Validation Data} & 10k images from CIFAR-10 \\
        \hline
        \textbf{Operation} & \textbf{Proc Time(h)} \\
        \hline
        \textbf{Re-training for 100 epochs} & $ 0.022 \times 100 = 2.276 $ \\
        \hdashline
        \textbf{Pruning Compensation} & $0.071 \times 1 = 0.071 $ \\
        \textbf{CaP} & $0.076 \times 1 = 0.076$ \\
        \textbf{Acc-drop evaluation} & $0.0011 \times 48 = 0.053$ \\
        \hline
    \end{tabular}
\end{table}

Replacing re-training with pruning compensation is the main source of efficiency in our method. In this section, we present experiments evidence to demonstrate the data and time efficiency of pruning compensation.

For the evaluation of data efficiency, we test how will the pruning result change with different training data budget in table \ref{tbl:prune_var_stat} on VGG-19 pre-trained on ImageNet. Apart from limited access to the original training data of ImageNet, we also test the feasibility of using another dataset, PASCAL VOC 2012, and even a data-free manner. This data-free manner means using randomly generated data under standard normal distribution for compensation. This data-free manner totally cancel the effect of data distribution on the correlation between channels, and the pruning is only affected by pre-trained parameters. Note that we still use a validation set sampled from ImageNet training set to guide structural search. 

The pruning results in table \ref{tbl:prune_var_stat} reveal that our method can perform stably given limited data accessibility. With only 5\% of ImageNet training data, the pruning result(FLOPs drop) is only slightly impaired. The finding on VOC is similar. Compensation is robust against a reasonable change in data distribution. Using a random generated dataset has a major damage(25\% performance decay) to the compensation. However, because our structural search method can effectively confine the accuracy drop to a certain limit, the data-free manner is still useful in special scenarios where no training data is available.

As to time efficiency, we report the processing time of our method in table \ref{tbl:prune_imagenet} and table \ref{tbl:prune_cifar10} as an overview. We break down the processing time of our method and make a comparison with the time for re-training in table \ref{tbl:runtime} .

Pruning compensation takes a large part of the total processing time(64\% for ResNet-50 on ImageNet and 35\% for VGG-16 on CIFAR-10), because the estimation step makes use of the whole training set. The estimation of covariance computes the outer product of the feature vector, which is memory-consuming. So the mini-batch size during compensation is set to 48 for ResNet-50 on ImageNet and 64 for VGG-16 on CIFAR-10. This makes pruning compensation 2 to 3 times slower than re-training for one epoch, but pruning compensation runs only once. 

The processing time of CaP is only relevant to the kernel size and channel number of layers. So CaP occupies only 8\% of the total processing time for ResNet-50 on ImageNet(large-scale), while 35\% of the total processing time is occupied by CaP for VGG-16 on CIFAR-10(small-scale).

Compared with re-training, our structural search method evaluates model accuracy on the validation set more frequently. For each convolutional/fully-connected layer in VGG16 or each residual block in ResNet-50, the accuracy-drop guided sparsity search takes 3 steps. So the time expended on validation is a significant part(27\% for ResNet-50 on ImageNet and 26\% for VGG-16 on CIFAR-10).
\subsection{Evaluating the Effectiveness of Compensation-aware Pruning}

\begin{figure}[t]
    \centering
    \includegraphics[width=2.5in]{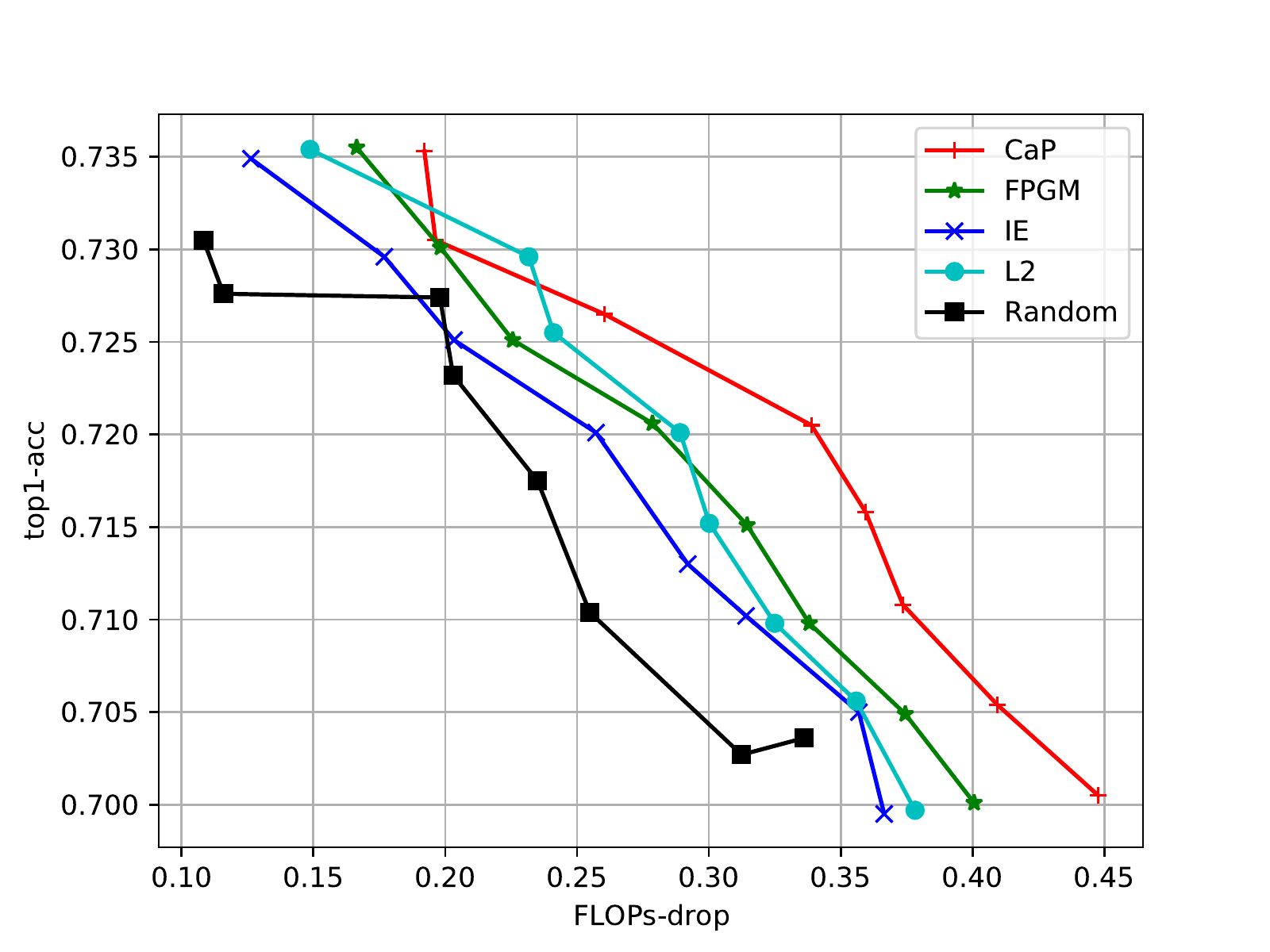}
    \caption{A comparison of CaP(ours), FPGM\cite{he2019filter}, IE(Importance Estimation)\cite{molchanov2019importance}, L2(L2-norm pruning), random pruning, using pruning compensation and structural search with step constraint. }
    \label{fig:var_pruning}
\end{figure}

\begin{figure}[t]
    \centering
    \includegraphics[width=3.5in]{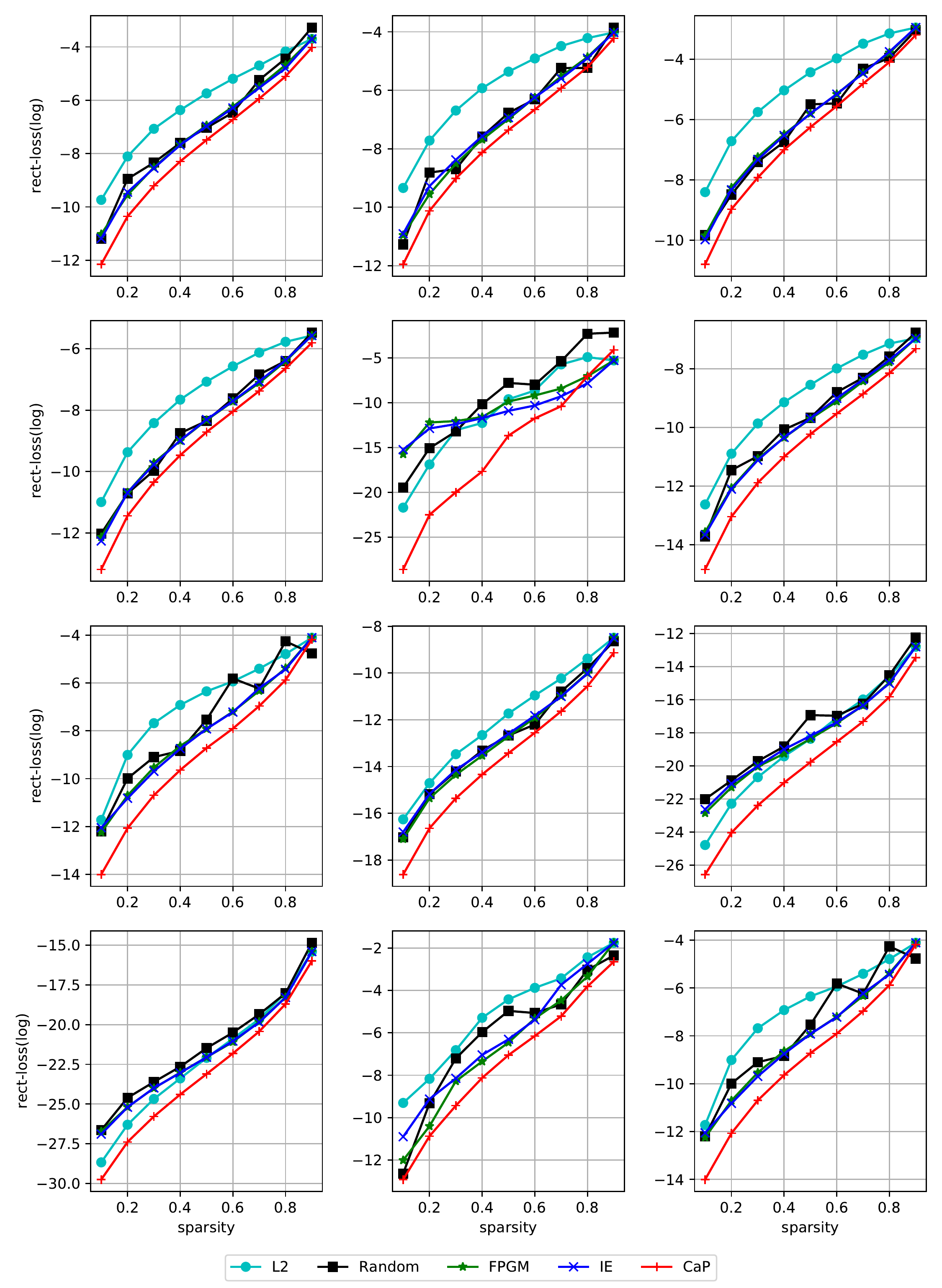}
    \caption{The growth of reconstruction loss(defined by equation \eqref{eq:CaP}, viewed in log-scale) at different sparsity rates with CaP and other counter-part pruning algorithms. The indexed layer in each sub-plot is from VGG-16 pre-trained on CIFAR-100. }
    \label{fig:rect_loss}
\end{figure}

We propose CaP as a novel heuristic-based pruning method in place of other alternatives, because we believe that CaP will have the best performance when working in collaboration with pruning compensation. We present experimental results in this sub-section to demonstrate the advantage of CaP.

We employ IE, FPGM, L2-norm, and a random pruning strategy to as the counter-parts against CaP. The model in the structural search is pruned by the chosen pruning algorithm(CaP or one of the counter-parts) and then compensated. The experiment is performed on VGG-19 pre-trained on CIFAR-100.

Figure \ref{fig:var_pruning} shows the pruning results. CaP shows superior performance over other pruning algorithms. Apart from random pruning who shows unstable performance, the counter-part pruning algorithms do not have distinct advantage over each other, while CaP has better performance in most cases.

An in-depth analysis on each pruning algorithm is presented in figure \ref{fig:rect_loss}. The aim of pruning algorithm in our method is to minimize the reconstruction loss in equation \eqref{eq:CaP}. We visualize and compare the reconstruction loss of the channel selection decision made by each algorithm at varying sparsity rate in figure \ref{fig:rect_loss}. Though avoidably the reconstruction loss will approach the maximum when the sparsity rate is close to 1, CaP preserves more channel information and has the least reconstruction loss. Meanwhile, we find that L2-norm pruning is even weaker than random pruning. It's unsurprising when we consider the fact that L2-norm pruning only take weights into account and ignore channel information. In a BN-enabled network, the magnitude of kernels is not a good indicator of channel importance or variance, because BN will rescale the channel output to a standard distribution after convolution.

\subsection{Analysis on Structural Search Results}
\label{subsec:search}
\begin{figure}[!t]
    \centering
    \includegraphics[width=3.5in]{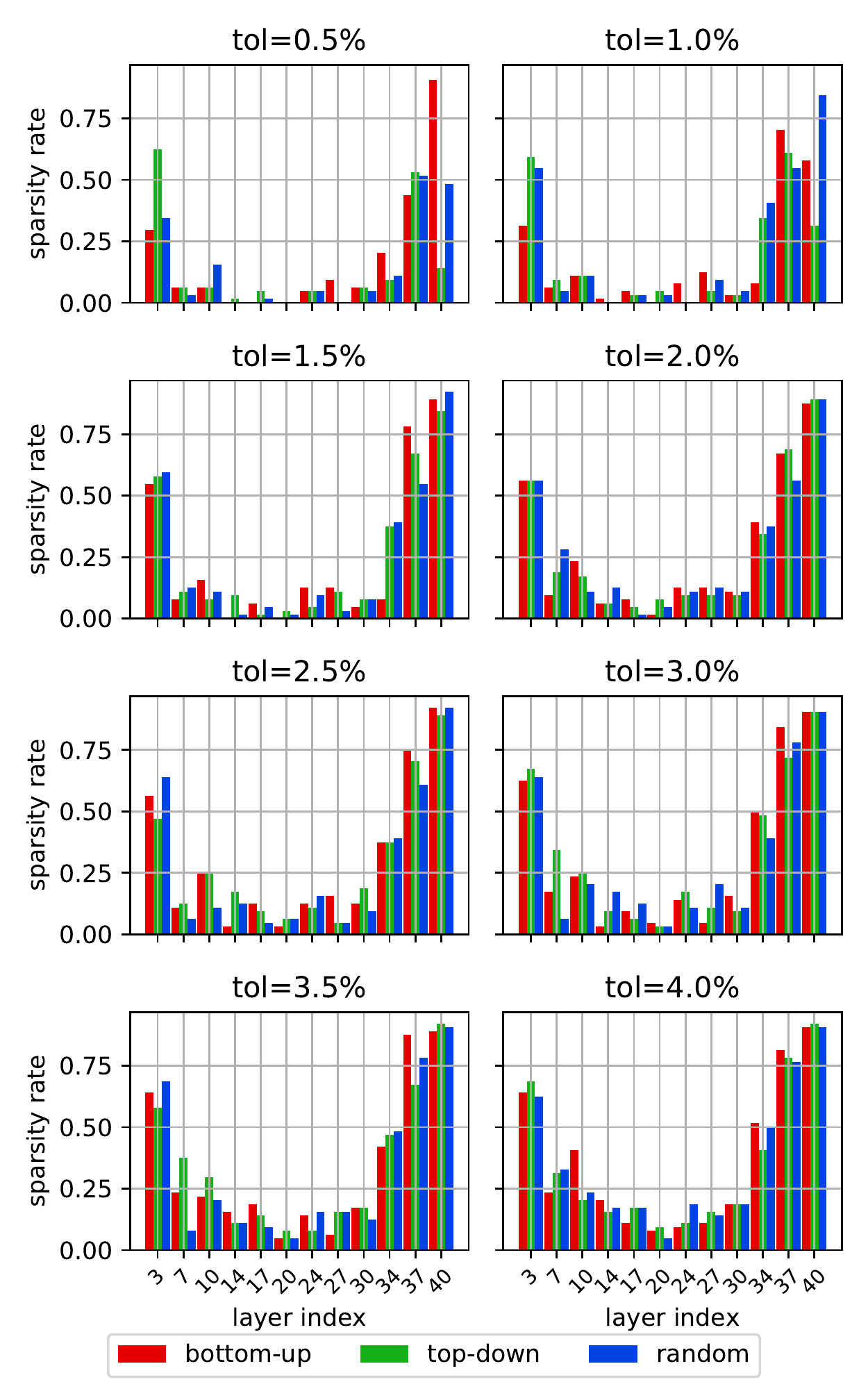}
    \caption{A comparison of search result(sparsity distribution) on convolutional layers of pruned VGG-16 pre-trained on CIFAR-100 by bottom-up, top-down, and random order.}
    \label{fig:order_comparison}
\end{figure}

\begin{figure}[!t]
    \centering
    \includegraphics[width=3.5in]{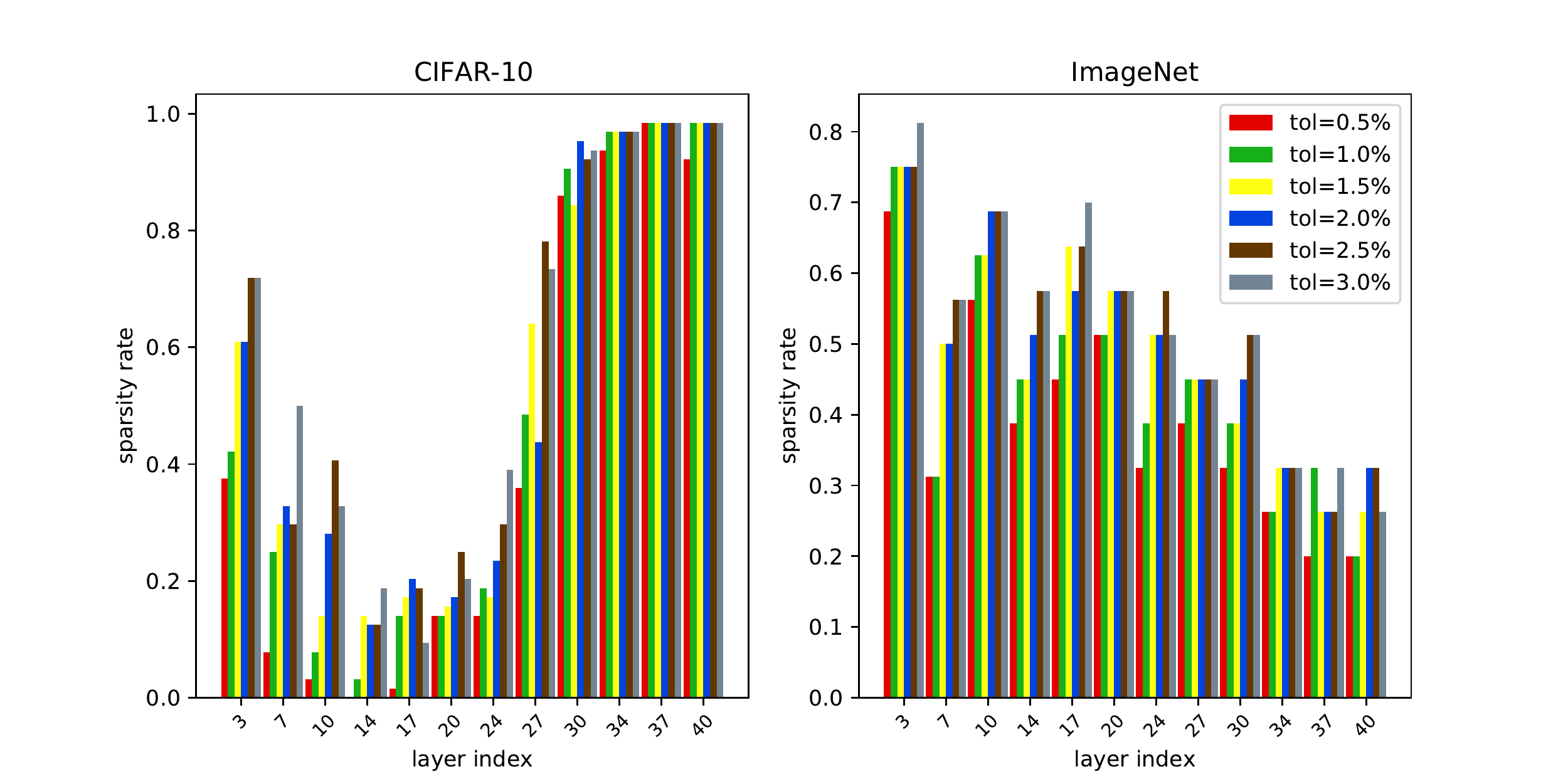}
    \caption{A comparison of search result(sparsity distribution) on convolutional layers of pruned VGG-16 pre-trained on CIFAR-10(left) and ImageNet(right).}
    \label{fig:sparsity_comparison}
\end{figure}

\begin{table}[!t]
    \renewcommand{\arraystretch}{1.3}
    \caption{FLOPs drop of VGG-19 pre-trained on CIFAR-100 with different structural search order and tolerance.}
    \label{tbl:prune_var_order}
    \centering
    \begin{tabular}{c || c | c | c | c }
    \hline
    \footnotesize{\textbf{Order}} & 
	\footnotesize{\textbf{tol=0.5\%}} & 
	\footnotesize{\textbf{tol=1.0\%}} & 
	\footnotesize{\textbf{tol=1.5\%}} & 
	\footnotesize{\textbf{tol=2.0\%}} \\

    \hline
    Bottom-up & 19.21 & 19.63 & 26.04 & 33.89 \\
    Top-down & 18.10 & 23.05 & 28.96 & 32.88 \\
    Random & 16.48 & 25.62 & 28.45 & 33.82 \\

    \hline
    & 
	\footnotesize{\textbf{tol=2.5\%}} & 
	\footnotesize{\textbf{tol=3.0\%}} & 
	\footnotesize{\textbf{tol=3.5\%}} & 
	\footnotesize{\textbf{tol=4.0\%}} \\

    \hline
    Bottom-up & 35.95 & 37.36 & 40.94 & 44.77 \\
    Top-down & 36.20 & 39.64 & 42.65 & 42.47 \\
    Random & 32.96 & 38.10 & 39.01 & 44.72 \\
    \hline

    \end{tabular}
\end{table}

We analyze the results of binary structural search with step constraint in this sub-section. We have two key findings about structural search from experiments.

First, the visiting order of layers has important but unpredictable relation with the pruning result. We test three visiting orders: bottom-up, top-down and random. A bottom-up order is the same as the visiting order of model inference, i.e. the breath first search order on the computational graph of the model. Top-down order is the reverse of bottom-up order. Pruning results in table \ref{tbl:prune_var_order} suggest that the visiting order makes a difference at a given tolerance constraint. For example, when the tolerance is $1\%$, the FLOPs drop of random order is $5.99\%$ higher than bottom-up order. However, none of the orders is dominantly superior over others or has stably good performance.

For more insight, we plot the final sparsity rate of each layer found by the three orders in figure \ref{fig:order_comparison}. By comparing the found structure under different tolerance, We can find the difference made by the search order is smaller when the tolerance becomes larger. When tolerance is $4\%$, the result structures are very similar, in contrast to $0.5\%$ tolerance, where the result still shows over-pruning on early visited layers even if the step constraint takes effect. An explanation for this result is that there's a redundancy bottleneck in the model, which allow a certain degree of lossless pruning. Therefore, when tolerance is small, over-pruning occurs on early visited layers because pruning redundant channels will not penalize the accuracy. When the tolerance is large, the redundancy bottleneck will be broken and the price of pruning is always paid. The sparsity limit of a layer is pushed by the step constraint only, regardless of the visiting order.

Our second finding is that the training data gives strong prior information about the optimal pruned structure, while baseline model structure, accuracy drop tolerance and pruning order are not very informative.

In figure \ref{fig:order_comparison}, though the layer sparsity increases evenly when the tolerance becomes larger and the pruning order causes perturbation on the sparsity, the final structure of the pruned models share a similar U-shape structure. This observation indicates that the redundancy of channels is determined by the training data, irrelevant to the post-hoc pruning algorithm or structural search. In figure \ref{fig:sparsity_comparison}, we plot the found structure of VGG-16 pre-trained on CIFAR-10 and ImageNet with different tolerance. CIFAR-10 has only 10 classes and $32\times 32$ input resolution, while ImageNet has 1000 classes and the input resolution is $224\times 224$. The final structures of the same model pre-trained on the two datasets are very different. For CIFAR-10 pre-trained VGG-16, most of the high-level layers are redundant, because 10-class classification does not need too much feature diversity. Meanwhile, most of the middle-level features are preserved, which still leads to a U-shape layer sparsity distribution. For ImageNet pre-trained VGG-16, the sparsity is more uniformly distributed among layers. The last two layers are less pruned, because the feature diversity at the high level is critical to the discriminability of the 1k-class classifier. This effect can also be found in other works(\cite{ding2020lossless, ruan2020edp, lin2021exemplar}), where the reported layer sparsity of the pruned model is similar to ours. Empirically, our finding is that the input dimension(e.g. image resolution for DCNN), data quantity and output dimension(e.g. class number) have direct influence on the optimal structure of models.

\section{Conclusions}

We propose a very efficient channel pruning method for DCNN that significantly reduces computational cost, data dependency, and human interference. Pruning compensation is proposed and verified experimentally to be an efficient and effective substitute of re-training for accuracy recovery. We also introduce a new pruning algorithm, compensation-aware pruning, to better preserve useful channels and take the advantage of pruning compensation. We completely automate the pruning procedure by performing a binary structural search with step constraint to prevent over-pruning. Through experimental evaluations, our method shows competitive pruning performance on FLOPs drop and accuracy preservation as the state-of-the-art pruning methods, meanwhile saving tens of GPU hours and a large amount of training data.

\appendices
\section{Efficient Computation of Compensation-aware Pruning}
\label{sec:appendix_a}

In CaP, the key computational complexity lies in the iterative computation of $\mathbf{\Sigma}_{\hat{S}, \hat{S}}^{-1}$, where $\mathbf{\Sigma}_{S, S}^{-1}$ has already been computed in the last iteration and $i$ is a tentative channel to join the channel selection($\hat{S}=S\cup \{i\}$). For each iteration, there are $|C-S|$ times of matrix inversion, which dramatically slow down the processing. We present and implement an efficient solution for this iterative augmented matrix inversion problem.

As the sample covariance matrix $\mathbf{\Sigma}_{\hat{S}, \hat{S}}$ and $\mathbf{\Sigma}_{S, S}$ are ensured to be positive-definite, they can be decomposed by Cholesky factorization:

\begin{align}
    & \mathbf{\Sigma}_{\hat{S}, \hat{S}} = \textbf{L}_{\hat{S}}\textbf{L}_{\hat{S}}^T \quad \mathbf{\Sigma}_{S, S} = \textbf{L}_{S}\textbf{L}_{S}^T
\end{align}
where $\textbf{L}_{S}\in\mathbb{R}^{|S|\times |S|}$ and $\textbf{L}_{\hat{S}}\in\mathbb{R}^{(|S|+1)\times(|S|+1)}$ are real lower triangular matrix with positive diagonal entries. An efficient way to compute the inverse of the symmetric real-valued positive-definite covariance matrix via Cholesky factorization is:

\begin{align}
    & \mathbf{\Sigma}_{\hat{S}, \hat{S}}^{-1} = (\textbf{L}_{\hat{S}}^T)^{-1}\textbf{L}_{\hat{S}}^{-1} \quad \mathbf{\Sigma}_{{S}, {S}}^{-1} = (\textbf{L}_{S}^T)^{-1}\textbf{L}_{S}^{-1}
\end{align}
where computing the inverse of a triangular matrix, i.e. $(\textbf{L}_{\hat{S}}^T)^{-1}$ and $\textbf{L}_{\hat{S}}^{-1}$, will be significantly faster than that of $\mathbf{\Sigma}_{\hat{S}, \hat{S}}^{-1}$. But Cholesky factorization of a large matrix is still computationally expensive. Our key finding to reduce the cost is that $\textbf{L}_{\hat{S}}^{-1}$ can be constructed using $\textbf{L}_{S}^{-1}$, as:

\begin{align}
    \textbf{L}_{\hat{S}}^{-1} = \left[\begin{array}{@{}c}
  \begin{matrix}
    \textbf{L}_{S}^{-1} & 0_{|S|\times 1} \\
  \mathbf{r}_{1\times |S|}  & a_{1\times 1}
  \end{matrix}
\end{array}\right]\quad a>0
\end{align}

The inverse the constructed $\textbf{L}_{\hat{S}}^{-1}$ can also be easily constructed:

\begin{align}
    \textbf{L}_{\hat{S}} = \left[\begin{array}{@{}c}
    \begin{matrix}
    \textbf{L}_{S} & 0_{|S|\times 1} \\
    -\frac{\mathbf{r}^T \textbf{L}_S}{a}  & \frac{1}{a}
    \end{matrix}
  \end{array}\right]
\end{align}

By recovering the process of decomposing the covariance matrix by Cholesky factorization, we have:

\begin{align}
\textbf{L}_{\hat{S}}\textbf{L}_{\hat{S}}^T = 
\left[\begin{array}{@{}c}
    \begin{matrix}
        \mathbf{\Sigma}_{S,S} & -\frac{\mathbf{\Sigma}_{S,S}\mathbf{r}}{a} \\
    -\frac{\mathbf{r}^T\mathbf{\Sigma}_{S,S}}{a} & \frac{\mathbf{r}^T\mathbf{\Sigma}_{S,S}\mathbf{r} + 1}{a^2}
    \end{matrix}
\end{array}\right]
=\left[\begin{array}{@{}c}
    \begin{matrix}
        \mathbf{\Sigma}_{S,S} & \mathbf{\Sigma}_{S,i} \\
        \mathbf{\Sigma}_{i,S} & \mathbf{\Sigma}_{i,i} \\
    \end{matrix}
\end{array}\right]
\end{align}

The solution that satisfies this construction rule is:

\begin{align}
    & a = \frac{1}{\sqrt{\mathbf{\Sigma}_{i,i} - \mathbf{\Sigma}_{i,S}\mathbf{\Sigma}_{S,S}^{-1}\mathbf{\Sigma}_{S,i}}} \nonumber\\
    & \mathbf{r} = -a\mathbf{\Sigma}_{i,S}\mathbf{\Sigma}_{S,S}^{-1}\nonumber\\
\end{align}

The original sub-optimization problem in algorithm \ref{alg:cap} can therefore be simplified to:
\begin{align}
    & \arg\min_i \sum_{k=1}^{N}(\textbf{W}_k^T\mathbf{\Sigma}_{C, C}\textbf{W}_k - \textbf{W}_k^T\mathbf{\Sigma}_{C,\hat{S}}\mathbf{\Sigma}_{\hat{S},\hat{S}}^{-1}\mathbf{\Sigma}_{\hat{S},C}\textbf{W}_k) \nonumber\\
    \equiv & \arg\max_i \sum_{k=1}^{N}||\textbf{W}_k^T\mathbf{\Sigma}_{C, \hat{S}}(\textbf{L}_{\hat{S}}^{-1})^T||_2^2 \nonumber\\
    \equiv & \arg\max_i \sum_{k=1}^{N}(\textbf{W}_k^T\mathbf{\Sigma}_{C,S}\mathbf{r}^T + a\textbf{W}_k^T\mathbf{\Sigma}_{C,i})^2 \nonumber\\
\end{align}

\ifCLASSOPTIONcaptionsoff
  \newpage
\fi

\bibliographystyle{IEEEtran}
\bibliography{reference}
\end{document}